\documentclass[runningheads]{llncs}
\usepackage{subcaption}

\usepackage[T1]{fontenc}
\usepackage{graphicx,verbatim}
\usepackage {xcolor}
\usepackage[normalem]{ulem}
\usepackage{amsmath}
\usepackage{amsfonts}

\graphicspath{ {./images/} }

\begin{document}
\title{A Statistical 3D Stomach Shape Model for Anatomical Analysis}

\author{Erez Posner\inst{1}\orcidID{0000-0003-2778-1612} \and
Ore Shtalrid\inst{1} \and
Oded Erell \inst{1} \and
Daniel Noy \inst{1} \and
Moshe Bouhnik\inst{1}\orcidID{0000-0003-3003-2604}}
\authorrunning{E. Posner et al.}
%
\institute{Intuitive Surgical, Inc. \\ 1020 Kifer Road, Sunnyvale, CA \\
\email{\{erez.posner, ore.shtalrid, oded.erell ,daniel.noy,moshe.bouhnik\}@intusurg.com}}


\maketitle              
\begin{abstract}
Realistic and parameterized 3D models of human anatomy have become invaluable in research, diagnostics, and surgical planning. However, the development of detailed models for internal organs, such as the stomach, has been limited by data availability and methodological challenges. In this paper, we propose a novel pipeline for the generation of synthetic 3D stomach models, enabling the creation of anatomically diverse morphologies informed by established studies on stomach shape variability. Using this pipeline, we construct a dataset of synthetic stomachs. Building on this dataset, we develop a 3D statistical shape model of the stomach, trained to capture natural anatomical variability in a low-dimensional shape space. The model is further refined using CT meshes derived from publicly available datasets through a semi-supervised alignment process, enhancing its ability to generalize to unseen anatomical variations. We evaluated the model on a held-out test set of real stomach CT scans, demonstrating robust generalization and fit accuracy. We make the statistical shape model along with the synthetic dataset publicly available on 
GitLab\footnote{\label{github_repo}\texttt{https://gitlab.com/erez.posner/stomach\_pytorch}} to facilitate further research.
This work introduces the first statistical 3D shape model of the stomach, with applications ranging from surgical simulation and pre-operative planning to medical education and computational modeling. By combining synthetic data generation, parametric modeling, and real-world validation, our approach represents a significant advancement in organ modeling and opens new possibilities for personalized healthcare solutions.


\end{abstract}

\section{Introduction}

Accurate modeling of anatomical structures is essential in modern medicine, supporting advances in diagnostics, surgical planning, image-guided interventions, and personalized treatment design~\cite{fadero2014three}. As biomedical imaging techniques have improved in resolution and accessibility, there is an increasing demand for computational tools that can represent anatomical variability in a principled and quantitative manner.

Statistical Shape Models (SSMs) have emerged as a foundational framework for this purpose~\cite{Goparaju2022Benchmarking}. An SSM captures the geometric variability of a structure by learning a low-dimensional shape space from a population of anatomically corresponding samples. By encoding the statistical distribution of shape deformations, SSMs provide a generative model capable of representing anatomical variations. This generalization capacity makes them highly valuable in a variety of clinical tasks, such as shape analysis, pathology detection, and surgical planning~\cite{HEIMANN2009543,LINDNER20173}. From a technical point of view, SSMs align anatomical shapes across a population to compute a mean shape and characterize common patterns of variation. These models allow for operations such as interpolation and statistical analysis while preserving anatomical plausibility~\cite{rajamani2007statistical}.

In practice, SSMs have supported a range of biomedical applications. In the domain of preoperative and intraoperative surgical planning, SSMs can assist in planning surgical trajectories. For example, in~\cite{Cai2024}, an SSM was aligned with non-segmented patient-specific CT scans to automate trajectory planning for pelvic surgeries. Similarly, in intraoperative procedures, SSMs are utilized for visual navigation. As demonstrated in~\cite{alimohamadi2020spinal}, low-resolution CT and magnetic resonance images were used to create patient-specific SSMs for visual navigation in needle-based back pain surgeries.

SSMs are also critical in anatomical morphology analysis (AMA)\cite{WYSIADECKI2024100284}, which includes evaluating shape abnormalities, identifying potentially cancerous regions~\cite{Demerath1009_Morphometric_lesion}, predicting biomechanical disorders, and guiding implant design based on patient-specific anatomy~\cite{Boutillon_2022_Morphometric_SSM}.

Although the use of SSM in the applications mentioned has been widely adopted in bony structures such as the skull, femur, and spine~\cite{patil2023accuracy}, as well as in soft tissue organs such as the liver and heart~\cite{rodero2021linking}, the stomach has received comparatively little attention in the literature. Yet, accurate modeling of the stomach is of significant clinical importance. The stomach is a highly deformable, hollow organ with substantial anatomical variability across individuals and dynamic shape changes due to food intake, posture, and pathology~\cite{Kim2015}. These characteristics pose challenges for image segmentation and shape analysis, especially in the context of bariatric surgery, where precise morphological understanding is critical for safe and effective intervention~\cite{salmaso2020computational}. Bariatric procedures such as sleeve gastrectomy or gastric bypass require detailed knowledge of stomach curvature, wall thickness, and volume to minimize complications and optimize postoperative outcomes~\cite{Robb2022}. Moreover, shape changes to the stomach are also relevant in other surgical contexts, including tumor resection, gastrectomy for malignancy, and reconstruction procedures~\cite{marsh2024gastric}.

The absence of a dedicated statistical shape model for the stomach also limits the ability to perform population-based morphometric analysis, quantify post-surgical morphological changes, or simulate plausible anatomical variations for training and planning purposes. Without such a framework, it is difficult to establish normative shape distributions, identify shape outliers, or robustly inform clinical decision-making in bariatric and oncological contexts~\cite{Boutillon_2022_Morphometric_SSM}.

To address these limitations, we propose a dedicated statistical shape modeling framework for the human stomach that integrates both synthetic and real-world data. Our approach is designed to capture anatomical diversity, support clinical interpretability, and generalize across imaging modalities and patient populations.

We introduce a 3D Statistical Shape Model (SSM) pipeline tailored to the stomach's complex morphology. The pipeline begins with the generation of a synthetic dataset composed of anatomically diverse stomach shapes, constructed using prior anatomical knowledge and procedural modeling techniques~\cite{Karnul2022-fq}. This synthetic dataset enables the learning of a low-dimensional shape space that encodes the principal modes of stomach shape variation. To ensure clinical applicability, the model is refined using a semi-supervised alignment strategy on real patient CT scans, enabling it to fit previously unseen morphologies while preserving anatomical plausibility~\cite{Cai2024}. This hybrid approach overcomes data scarcity and annotation variability by leveraging synthetic priors while adapting to real-world variability.

Our contributions are threefold:

\textbf{Synthetic Data Generation}: We develop a novel pipeline to generate anatomically diverse synthetic stomach morphologies, creating a dataset of stomach models based on established anatomical studies.

\textbf{Statistical shape space creation}: Using the synthetic dataset, we train a low-dimensional shape model that captures the variability of stomach shape.

\textbf{Real-World Refinement}: The model is refined through semi-supervised alignment with real CT scans, ensuring its applicability to real-world data and unseen morphologies. We evaluated the model on a test set of real stomach CT scans, demonstrating its ability to generalize and accurately fit diverse anatomical variations.

\section{Method}
We propose a three-stage framework to develop a statistical 3D shape model of the human stomach. First, we generate a synthetic dataset of stomach morphologies using a novel parametric stomach generation pipeline (Sec.~\ref{sec:dataset}). Next, we train a low-dimensional model of stomach shapes using this synthetic dataset (Sec.~\ref{sec:learning_synthetic_data}). Finally, we refine the shape model by fitting it to real-world CT scans, improving its ability to generalize to unseen anatomies (Sec.~\ref{refinement_with_real_scans}).
\subsection{Dataset}
\label{sec:dataset}
We created a dataset consisting of 6,000 stomach models, each capturing distinct anatomical features and variations in adults. To construct the dataset, we first generated a template mesh of the stomach and then leveraged existing research on stomach morphology~\cite{Karnul2022-fq} to compile a comprehensive set of anatomical characteristics. An automatic pipeline for stomach generation was implemented to ensure the diversity and representativeness of the dataset.


\subsubsection{Stomach 3D template mesh}
\begin{figure}[tbp!]
    \centering
    \begin{tabular}[t]{cc}

            \begin{subfigure}{.33\textwidth}
            \begin{tabular}{c}
                \smallskip

                    \begin{subfigure}[t]{1\textwidth}
                        \centering
                        \includegraphics[width=1\textwidth]{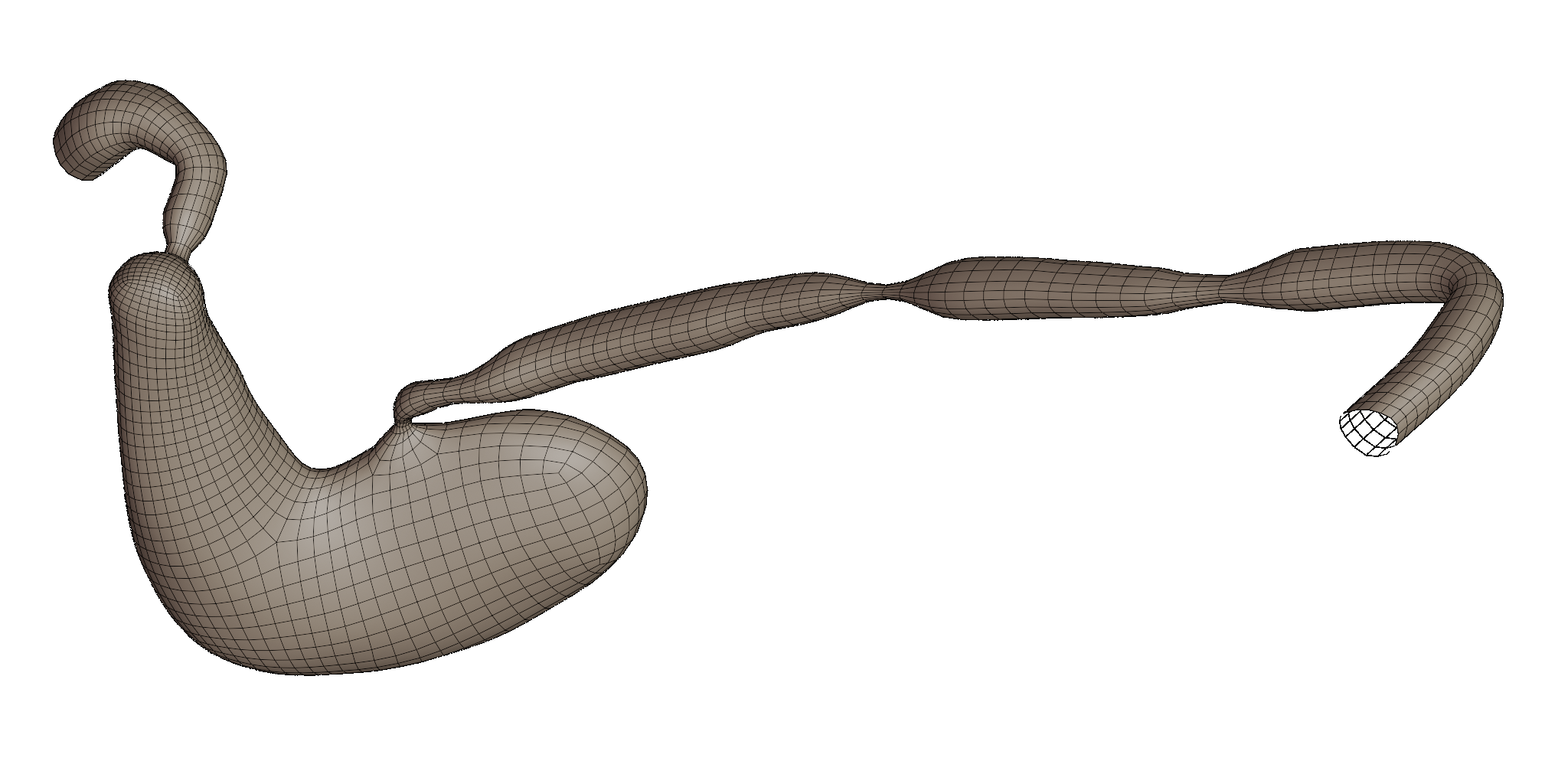}
                    \end{subfigure}
            \end{tabular}
                    \end{subfigure}

        \begin{subfigure}{.33\textwidth}
            \begin{tabular}{c}
                \smallskip
 
                    \begin{subfigure}[t]{1\textwidth}
                        \centering
                        \includegraphics[width=1\textwidth]{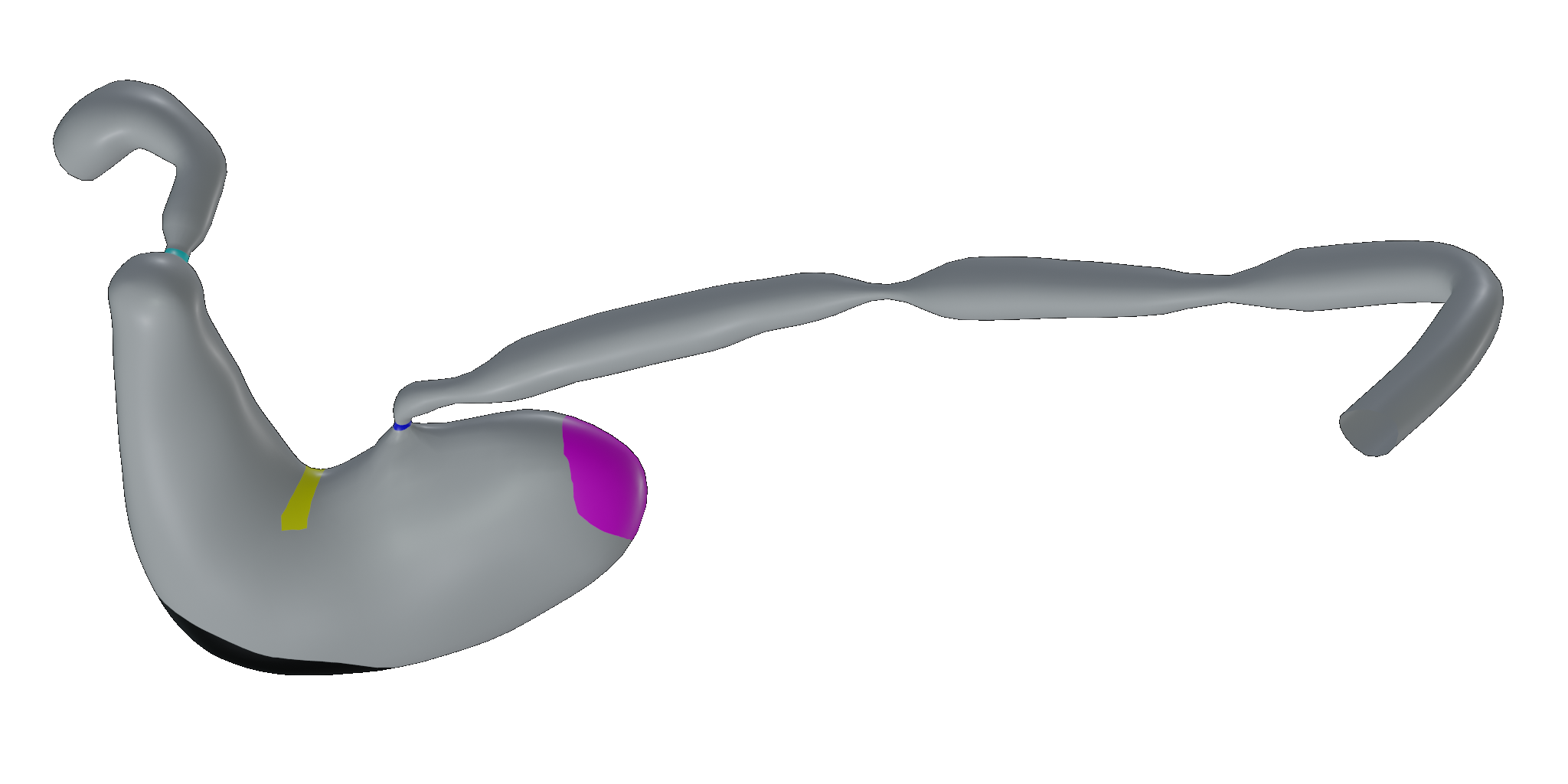}
                    \end{subfigure}
            \end{tabular}
        \end{subfigure}

        \begin{subfigure}{.33\textwidth}
            \begin{tabular}{c}
                \smallskip
 
                    \begin{subfigure}[t]{1\textwidth}
                        \centering
                        \includegraphics[width=1\textwidth]{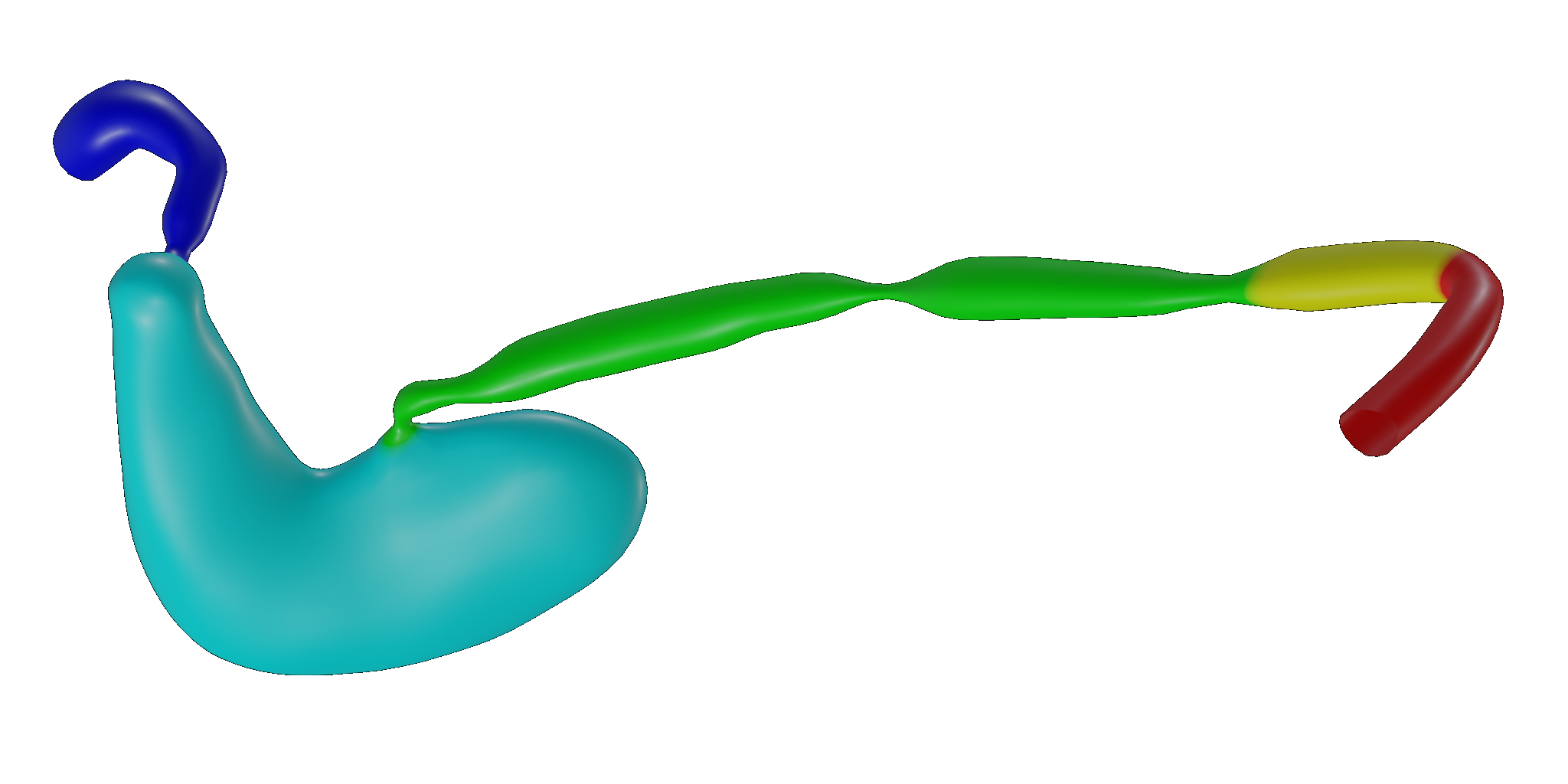}
                    \end{subfigure}
            \end{tabular}
        \end{subfigure}

    \end{tabular}

    \caption{Three mesh representations of the template are shown. Left: The average template mesh. Middle: The template mesh with anatomical landmarks annotated using color-coded labels: incisura angularis [yellow], greater-curvature [black], esophageal-junction [blue], pyloric-sphincter [cyan], and fundus [purple]. Right: The template mesh with anatomical regions annotated using color-coded labels: mouth [red], pharynx [yellow], esophagus [green], stomach [cyan], and duodenum [blue].} \label{template_mesh_viz}
\end{figure}

The 3D template mesh of the stomach, which serves as the foundation of our modeling pipeline, was derived from 170 real-world stomach CT scans in the \textit{WORD} dataset~\cite{WORD_DATASET}. First, each segmented stomach was converted into a triangular mesh using the Marching Cubes algorithm~\cite{marching_cubes}. The meshes were then aligned to the Gastroesophageal Junction (GEJ). Next, we created volume-rendered~\cite{volume_rendering} visualizations along the coronal, sagittal, and transverse planes (See \texttt{template\_mesh.mp4} in Supplementary) and used them as a visual guide for an iterative, hand-sculpting process. Specifically, by examining the volume renderings, we manually sculpted a shape that represents an approximate average stomach geometry across all 170 aligned meshes. This ensured that the final template mesh is smooth and anatomically accurate.

The stomach template mesh consists of V=3,424 vertices and K=23 joints, featuring a multi-hierarchical part segmentation (e.g., fundus, body, pylorus) that was manually defined using standard anatomical references~\cite{landmarknames}. The template mesh was carefully reviewed by a gastroenterology expert to ensure clinical accuracy. The mesh includes a skeletal rig to facilitate smooth deformations and a consistent topology. Fig.~\ref{template_mesh_viz} shows the multi-hierarchy part segmentation for anatomical landmarks and regions. This standardized template captures the stomach's structural complexity and provides a strong foundation for modeling its diverse morphologies.

\subsubsection{Stomach Morphology Modeling} \label{stomach_morphology}
To generate realistic and anatomically plausible stomach 3D meshes, we designed a pipeline informed by the anatomical variations studied in the work of Karnul et al~\cite{Karnul2022-fq}. The pipeline applies four distinct groups of operations to the stomach template mesh, enabling the creation of morphologically diverse stomachs. The first group focuses on the dimensional parameters, including Greater Curvature length (GC), Lesser Curvature length (LC), and the stomach volume. These attributes provide precise control over measurable features, ensuring that the generated meshes fall within realistic anatomical ranges. The second group models shape-type variability, leveraging the four primary structural categories—Cylindrical, J-shaped, Reverse L-shaped, and Crescentic—identified in the study by~\cite{Karnul2022-fq}. Each morphology is represented by an adjusted version of the template mesh, and an interpolation space is constructed to allow for blending between these types. This facilitates the generation of stomach shapes that capture nuanced variations, blending characteristics of multiple morphological types. The third group simulates surgical insufflation by expanding the stomach outward, displacing vertices along their normals while preserving anatomical accuracy and preventing self-collisions. The fourth group introduces procedural shape jittering, where noise-based deformations are applied to specific regions of the stomach. For example, the pylorus may be rotated by a random angle, or the fundus scaled relative to its rest pose, generating diverse and anatomically plausible shape variants.

These operations collectively ensure the generation of a dataset encompassing a broad range of stomach morphologies, reflecting real-world anatomical diversity. While anatomical parameters are varied independently, the variation is constrained to physiologically plausible ranges, following a domain randomization strategy. This helps maintain realistic anatomical relationships without requiring explicit modeling of interdependencies.

\subsubsection{Stomach Generation Pipeline} \label{subsec:hss}
The pipeline is implemented using a standard skinning scheme~\cite{10.1145/2010324.1964973}, allowing for smooth deformations of the stomach mesh via skeletal manipulations. The skeletal rig comprises control handlers corresponding to anatomical regions, with blend weights calculated using biharmonic functions~\cite{10.1145/2010324.1964973} over a constrained tetrahedral mesh derived from the stomach template. 
The pipeline applies operations sequentially:

\begin{itemize}
    \item Shape-type attributes and procedural jittering are implemented first. Morphologies are generated by interpolating among the four predominant shapes and applying controlled perturbations to handlers affecting specific anatomical regions (e.g., rotating the pylorus or scaling the fundus). 
    \item Surgical insufflation is then simulated by extruding vertices outward along their normal while ensuring collision-free deformations.
    \item Dimensional parameters, such as GC, LC, and stomach volume, are constrained last by scaling the mesh to achieve target values.
\end{itemize}
Although anatomical parameters are varied independently, the skeletal structure and biharmonic skinning ensure that deformations propagate smoothly across the mesh, preserving coherent anatomical relationships across regions. Furthermore, by explicitly constraining measurable anatomical attributes (e.g., GC, LC, volume) to physiologically plausible ranges, the pipeline implicitly maintains realistic covariation patterns without requiring an explicit statistical model. This enables the generation of diverse, anatomically plausible shapes that reflect both structural variation and preserving global anatomical coherence observed in real-world data. A detailed overview of parameter ranges for specific transformations and visualization of the pipeline are provided in GitLab\footref{github_repo} and Supplementary, respectively.

\subsubsection{Dataset Validation and Diversity}
The final dataset consists of 6,000 stomach meshes, uniformly sampled across key anatomical attributes described in Sec~\ref{stomach_morphology} To ensure physiological plausibility, we collaborated with a gastroenterology expert to refine the pipeline and eliminate anatomically implausible outputs. Each generated mesh undergoes automated validation, including self-collision detection and bounds checking on key anatomical parameters such as curvature lengths and overall volume. These tests ensure that all meshes remain free of geometric artifacts and fall within clinically realistic ranges.

\subsection{Statistical 3D Stomach Shape Model}
We learn the stomach statistical shape model in two stages. First, we create an initial shape model from the set of synthetic stomachs, referred to as prototypes. Then, we align the initial model, in a semi-unsupervised manner, to multiple 3D stomach scans from publicly available datasets~\cite{li2023medshapenetlargescaledataset,WORD_DATASET}. After alignment, we refine the parameters of the initial model. In the following, we describe the stomach model and the procedure for learning its parameters from prototypes and subsequent refinements using real CT scans.
\subsubsection{Learning from Synthetic Data}\label{sec:learning_synthetic_data}

The stomach shape model is designed to capture anatomical variability in a low-dimensional shape space. It represents stomach shapes as deformations of a standardized template mesh. The model is defined as the function $M(\beta, \theta, \gamma)$ that returns a 3D mesh $v \in \mathbb{R}^{3\times V}$, where:
\begin{itemize}
    \item $\beta \in \mathbb{R}^m$ is a vector of the coefficients of the learned shape space.
    \item $\theta$ is the orientation of the model, represented as a global rotation.
    \item $\gamma$  is the global translation applied to the model.
\end{itemize}
In details, the shape of the stomach is generated as follows:
\begin{equation}
v_s = v_t + B_s\beta^T
\end{equation}
where $v_t \in \mathbb{R}^{3\times V}$ is the template mesh, and $B_s \in \mathbb{R}^{3\times V \times m}$ is the learned basis matrix capturing $m$ principal shape directions. A rigid transformation is applied to account for pose and translation:
\begin{equation}
v = R(\theta)v_s + \gamma
\end{equation} 
This formulation captures the global anatomical variability of the stomach in a compact, low-dimensional space.
To create the initial shape model, we leverage the synthetic stomach meshes generated using our pipeline. These meshes share a consistent topology with the template mesh, making them suitable for learning a linear shape space using Principal Component Analysis (PCA).

PCA is applied to the vertex positions of the prototypes to calculate the mean shape $v_t$ and the principal components $B$, which capture the main modes of variation in the dataset. 

\subsubsection{Refinement with Real CT Scans}\label{refinement_with_real_scans}
To refine the initial stomach model learned from synthetic prototypes, we leverage two publicly available datasets~\cite{li2023medshapenetlargescaledataset,WORD_DATASET} containing real CT scans. The \textit{WORD}~\cite{WORD_DATASET} and MedShapeNet~\cite{li2023medshapenetlargescaledataset}  datasets provide access to a broad spectrum of real stomach anatomies. \textit{WORD}~\cite{WORD_DATASET} anatomies influenced by common pathologies. MedShapeNet~\cite{li2023medshapenetlargescaledataset} complements this with meshes from multiple institutions, covering a wider range of ages, body types, imaging protocols, and both healthy and diseased cases. Together, they enable realistic refinement across diverse anatomical and demographic variations.

In the \textit{WORD}~\cite{WORD_DATASET} dataset, each voxel of the CT volume is labeled by organ, allowing us to extract the stomach region and reconstruct it as a 3D mesh via the Marching Cubes algorithm~\cite{marching_cubes}. Similarly, MedShapeNet~\cite{li2023medshapenetlargescaledataset} offers ready-to-use stomach meshes. A clinical expert annotated the CT-derived stomach meshes with five anatomical landmarks~\cite{landmarknames}—the Gastroesophageal Junction, the fundus, the greater curvature, the Angular Incisure, and the pyloric sphincter — which are essential to guide the alignment process. 

The alignment is performed using co-registration~\cite{10.1007/978-3-642-33783-3_18}. First, we fit the initial shape model to the CT-derived stomach meshes. Then, we run a model-free step where we regularize the registration to the current model by adding a coupling term. To fit the shape model to scans, we minimize the objective:
\begin{equation}
E(\beta, \theta, \gamma, dv) = E_{data}(\beta, \theta, \gamma, dv) + E_s(\beta) + E_{coup}(v),
\end{equation}
where \(E_{data}\) captures the alignment between the mesh and the scan using symmetric point-to-point distances, normal consistency, and keypoint alignment at anatomical landmarks. It is defined as:

\begin{align}
E_{data}(\beta, \theta, \gamma, dv) = 
& \lambda_{p2p} \left( \sum_{p \in \mathcal{V}} \min_{q \in \mathcal{S}} \| p - q \|^2 
+ \sum_{q \in \mathcal{S}} \min_{p \in \mathcal{V}} \| q - p \|^2 \right) \notag \\
& + \lambda_{n} \sum_{p \in \mathcal{V}} \min_{q \in \mathcal{S}} | (p - q) \cdot \mathbf{n}_q | + \lambda_{lm} \sum_{l \in \mathcal{L}} \| l_{\text{model}} - l_{\text{scan}} \|^2,
\end{align}
where \(p \in \mathcal{V}\) are the mesh points,  \(q \in \mathcal{S}\) are scan points, \(\mathbf{n}_q\) are the normals associated with the scan points, and \(l_{\text{model}}\) and \(l_{\text{scan}}\) are the corresponding anatomical landmarks on the model and scan, respectively.

The shape prior term, $E_s(\beta)$ is represented as squared Mahalanobis distance, penalizing deviations from the initial stomach shape distribution and $E_{coup}(v)$ is a coupling term that regularizes the model-free registration to maintain alignment with the initial shape model fit.
The coupling term $E_{coup}(v)$  is defined as:
\begin{equation}
 E_{coup}(v)=\sum_{j=1}^{V}|v_i^0-(v_i+dv_i)|
\end{equation}
where $v_i^0$ is vertex $i$ of the initial model fit to the scan, and the $v_i+dv_i$ are the coupled mesh  vertices being optimized. This term ensures that the fitted mesh remains close to the initial shape model fit while adapting to the CT scan data.

\begin{figure}[thbp]
\centering
\includegraphics[width=0.6\textwidth]{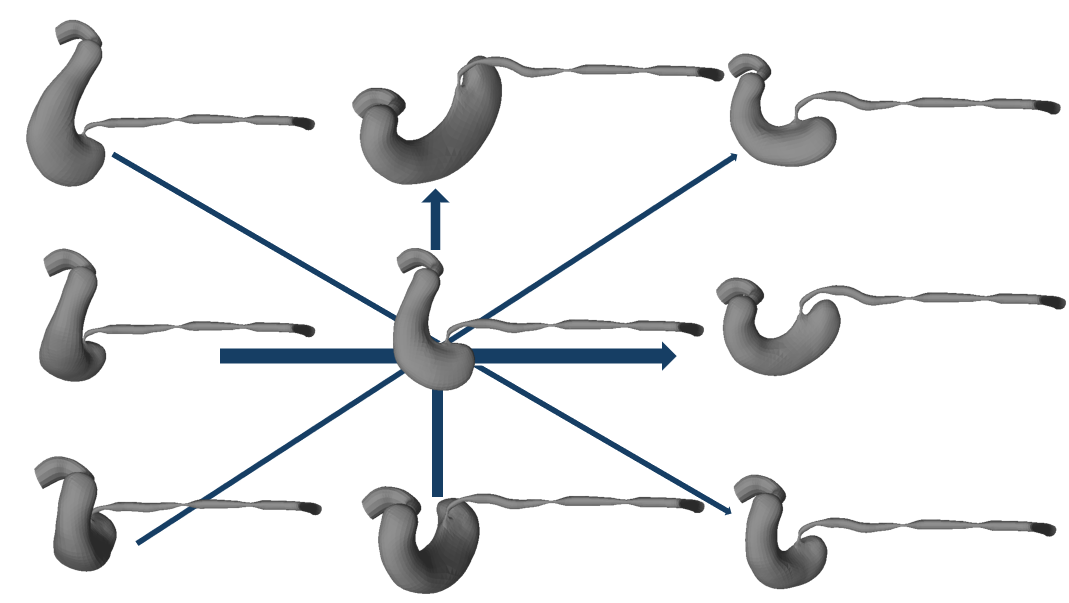}
\caption{First and second principal component varying from left to right and bottom to up accordingly with \(\pm 3std)\)}
\label{fig:shape_space_results}\end{figure}

Optimization of this objective is implemented using PyTorch3D~\cite{ravi2020pytorch3d}. An annealing schedule is applied during optimization, gradually reducing the weight of the shape prior and increasing the influence of the data term.

After fitting the shape model to the CT scans, we re-learn the shape space by applying PCA to the aligned meshes. To ensure a balanced representation, we combine the refined CT-based samples with a subset of synthetic samples, weighting them equally in the PCA computation. This updated shape space incorporates real-world anatomical variability while retaining the diversity captured by the synthetic data. The final principal components are visualized in Fig.~\ref{fig:shape_space_results}.

\begin{figure}[hbt!]
  \centering
    \begin{tabular}[t]{ccc}

        \begin{subfigure}{0.33\textwidth}
            \begin{tabular}{c}
                \smallskip

                    \begin{subfigure}[t]{1\textwidth}
                        \centering
                        \includegraphics[width=1\textwidth]{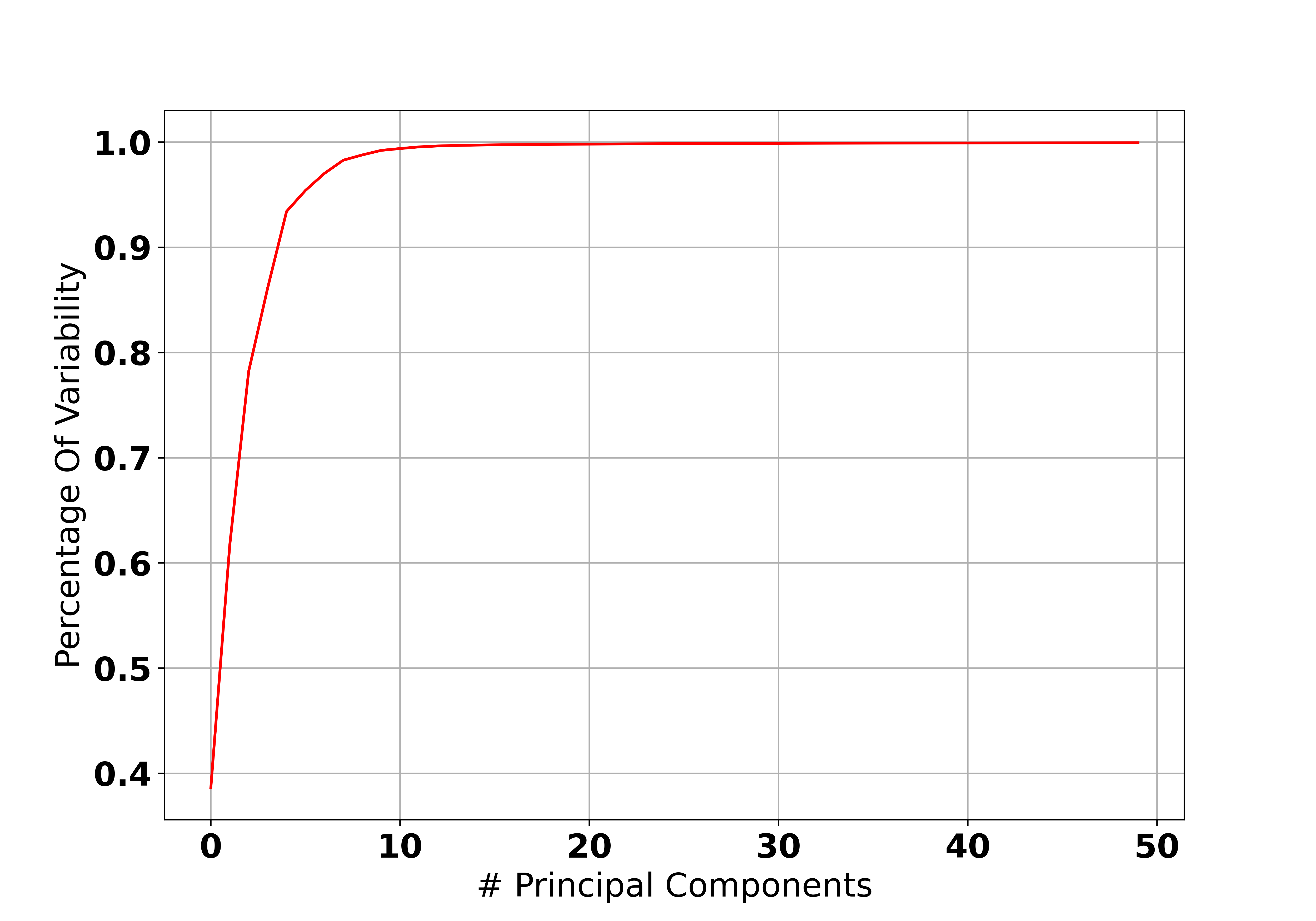}
                    \end{subfigure}
            \end{tabular}
                    \end{subfigure}

        \begin{subfigure}{0.33\textwidth}
            \begin{tabular}{c}
                \smallskip
 
                    \begin{subfigure}[t]{1\textwidth}
                        \centering
                        \includegraphics[width=1\textwidth]{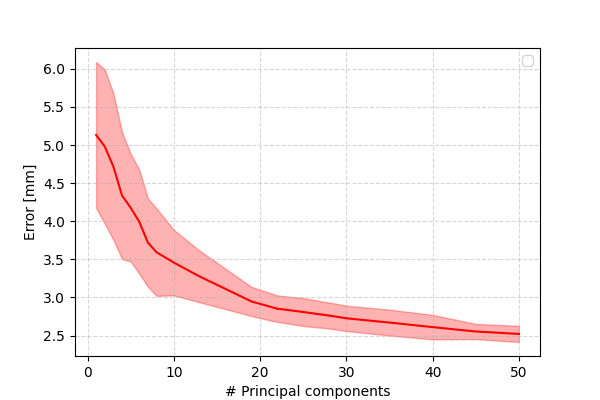}
                    \end{subfigure}
            \end{tabular}
        \end{subfigure}
        \begin{subfigure}{0.33\textwidth}
            \begin{tabular}{c}
                \smallskip
 
                    \begin{subfigure}[t]{1\textwidth}
                        \centering
                        \includegraphics[width=1\textwidth]{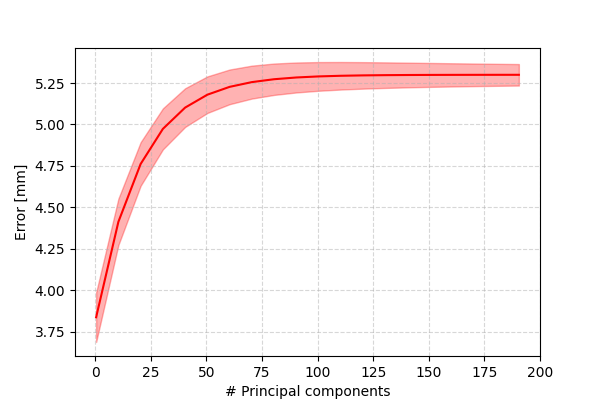}
                    \end{subfigure}
            \end{tabular}
        \end{subfigure}

    \end{tabular}
\caption{Quantitative evaluation of the shape model. The plots show compactness (variance explained on training data), generalization (mean fitting error on held-out test meshes), and specificity (average distance from random samples to nearest training shape)}
\label{fig:quantitative-results}
\end{figure}

\begin{figure}[hbt]
\centering
\includegraphics[width=1\linewidth]{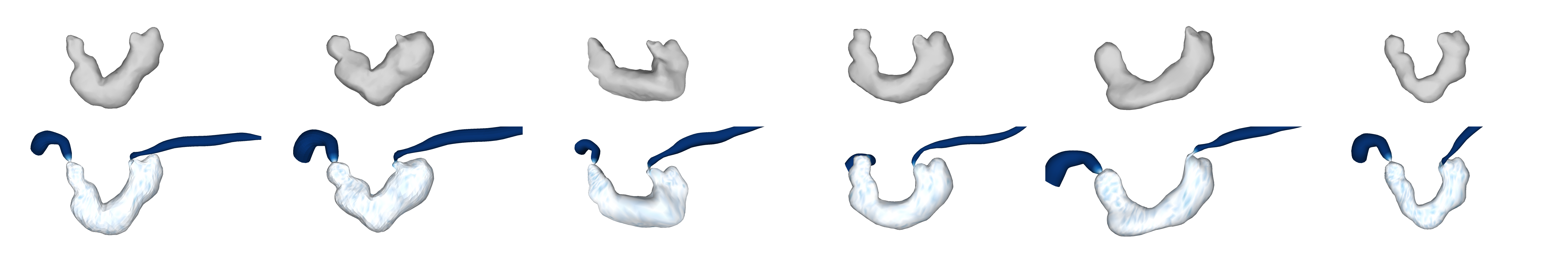}
\caption{Qualitative results of shape model fitting to test meshes. Each column shows (top to bottom): the ground truth CT-extracted mesh and the predicted shape model mesh, colored by per-vertex mesh-to-scan distance. Distances are clipped at 4 mm and normalized for visualization; darker blue regions indicate higher error.}
\label{fig:qualitative-results}\end{figure}

\section{Experiments}
A well-designed statistical shape model should be compact—capturing most of the data variability with as few parameters as possible—while also generalizing well to unseen examples and remaining specific to anatomically valid instances of the object class. A widely accepted way to quantify these properties is through three complementary metrics: compactness, generalization, and specificity~\cite{Karnul2022-fq}. These criteria have been used extensively to evaluate statistical models across various anatomical and non-anatomical domains, such as 3D face modeling~\cite{10.1145/3130800.3130813,7410768,10.1007/s11263-017-1009-7}. Together, they offer a principled framework for selecting a model dimensionality that balances expressiveness and robustness, without overfitting to training data.

We used 200 aligned CT-derived stomach meshes, randomly sampled from the \textit{WORD}~\cite{WORD_DATASET} and MedShapeNet~\cite{li2023medshapenetlargescaledataset} datasets in similar proportions, along with 6,000 synthetic prototypes, to learn the shape space. To account for the imbalance between real and synthetic data, we applied a weighted PCA formulation that emphasizes the contribution of real anatomies during shape space construction. Evaluation was conducted on a held-out test set of 100 stomach meshes, randomly sampled from MedShapeNet~\cite{li2023medshapenetlargescaledataset}. All test samples were excluded from the training and refinement processes to ensure an unbiased assessment.

The evaluation involved fitting the shape model to each test mesh using the energy function defined as:

\begin{equation}
E(\beta, \theta, \gamma) = E_{data}(\beta, \theta, \gamma) + E_s(\beta),
\end{equation}

where \(E_{data}(\beta, \theta, \gamma)\) and \(E_s(\beta)\) are defined as in Section~\ref{refinement_with_real_scans}. Unlike the refinement stage, the \(dv\) optimization term was removed, which means that only the shape space parameters (\(\beta, \theta, \gamma\)) were optimized. Similar to \cite{10.1145/3130800.3130813}, to evaluate the quality of our statistical shape model, we independently assessed its compactness, generalization, and specificity, as shown in Fig.~\ref{fig:quantitative-results}. The model was evaluated for a range of principal components, and all metrics were computed consistently across this range.

\textbf{Compactness} 
We measured compactness by computing the cumulative variance explained by the top 
$k$ principal components of the shape space. This metric reflects how efficiently the model captures variability in the training data. As shown in Fig.~\ref{fig:quantitative-results}, the model captures 98\% of the total variance with just 10 components, and effectively 100\% with 30 components. These results indicate that the learned shape space is low-dimensional yet expressive, enabling a compact representation of anatomical variation.

\textbf{Generalization} 
We quantify generalization by measuring how accurately the model can represent unseen anatomical shapes. For each mesh in the test set, we compute the average surface-to-surface distance between the fitted mesh and the corresponding ground truth. This provides a direct measure of the model’s ability to generalize beyond the training data. As shown in Fig.~\ref{fig:quantitative-results}, the generalization error decreases as the number of principal components increases. With 30 components, the mean mesh-to-scan distance remains below 2.7 mm. This result demonstrates that the model effectively captures the anatomical variability present in new samples. Additionally, the use of co-registration during preprocessing contributes to the robustness of the fits, particularly in the presence of scan noise and partial occlusions.

\textbf{Specificity} 
Specificity was evaluated by randomly sampling 1,000 shapes from the model’s latent space, reconstructing the corresponding meshes, and computing the average surface distance to the nearest mesh in the training set. This evaluation tests whether the model generates anatomically plausible shapes that remain consistent with the learned distribution. As shown in Fig.~\ref{fig:quantitative-results}, the specificity error remains low across different numbers of components, confirming that the model consistently produces realistic and anatomically valid stomach shapes.

As part of the qualitative evaluation, Figure~\ref{fig:qualitative-results} illustrates the model’s ability to reconstruct diverse stomach anatomies from the test set. Each example shows the ground-truth mesh extracted from CT data and the corresponding prediction generated by the shape model. Vertex-wise color maps visualize the reconstruction error, computed as the distance from each scan vertex to the closest point on the predicted surface. The examples demonstrate that the model captures both global shape and local anatomical details, even for previously unseen cases.


\section{Conclusions and Summary}
We presented a novel shape model of the human stomach, capturing anatomical variability in a compact shape space. Our approach combined synthetic data generation, refinement with real CT scans, and validation to ensure anatomical accuracy and robust generalization. By balancing synthetic and real-world samples, the final shape space reflects both simulated diversity and real anatomical features.
Quantitative and qualitative evaluations demonstrated accurate fits to test meshes and unseen morphologies. The publicly released model provides valuable tools for gastroenterology, surgical simulation, and medical imaging, enabling applications such as patient-specific planning, AI-driven diagnostics, and procedural training. Future work could extend this approach to other organs, explore the effect of human pose on stomach shape, and integrate more real-world data for improved fidelity.

%
%
%
\bibliographystyle{splncs04}
\bibliography{MICCAI2025_paper_template.bib}

\end{document}